\documentclass[conference]{IEEEtran}

\usepackage{cite}
\usepackage{amsmath,amssymb,amsfonts}
\usepackage{algorithmic}
\usepackage{graphicx}
\usepackage{textcomp}
\usepackage{xcolor}
\usepackage{tabularx}
\usepackage{url}
\usepackage{caption} 
\makeatletter
\newcommand{\linebreakand}{%
  \end{@IEEEauthorhalign}
  \hfill\mbox{}\par
  \mbox{}\hfill\begin{@IEEEauthorhalign}
}
\makeatother

\title{Supervised Anomaly Detection Method Combining Generative Adversarial Networks and Three-Dimensional Data in Vehicle Inspections}

\author{
\IEEEauthorblockN{Yohei Baba}
\IEEEauthorblockA{\textit{JR East Information Systems Company,}}
\IEEEauthorblockA{\textit{Shinjuku, Tokyo, 169-0072, Japan}}
\IEEEauthorblockA{\textit{ybaba@inet.jeis.co.jp}}
\and
\IEEEauthorblockN{Takuro Hoshi}
\IEEEauthorblockA{\textit{JR East Information Systems Company,}}
\IEEEauthorblockA{\textit{Shinjuku, Tokyo, 169-0072, Japan}}
\IEEEauthorblockA{\textit{houshi@inet.jeis.co.jp}}
\and
\IEEEauthorblockN{Ryosuke Mori}
\IEEEauthorblockA{\textit{East Japan Railway Company}}
\IEEEauthorblockA{\textit{Shibuya, Tokyo, 151-0053, Japan}}
\IEEEauthorblockA{\textit{r-mori@jreast.co.jp}}
\and
\linebreakand
\IEEEauthorblockN{Gaurang Gavai}
\IEEEauthorblockA{\textit{PARC A Xerox Company,3333 Coyote Hill Road, Palo Alto}}
\IEEEauthorblockA{\textit{CA 94304 USA}}
\IEEEauthorblockA{\textit{ggavai@parc.com}}

}

\begin{document}

\maketitle

\begin{abstract}
The external visual inspections of rolling stock’s underfloor equipment are currently being performed via human visual inspection. In this study, we attempt to partly automate visual inspection by investigating anomaly inspection algorithms that use image processing technology. As the railroad maintenance studies tend to have little anomaly data, unsupervised learning methods are usually preferred for anomaly detection; however, training cost and accuracy is still a challenge. Additionally, a researcher created anomalous images from normal images by adding noise, etc., but the anomalous targeted in this study is the rotation of piping cocks that was difficult to create using noise. Therefore, in this study, we propose a new method that uses style conversion via generative adversarial networks on three-dimensional computer graphics and imitates anomaly images to apply anomaly detection based on supervised learning. The geometry-consistent style conversion model was used to convert the image, and because of this the color and texture of the image were successfully made to imitate the real image while maintaining the anomalous shape. Using the generated anomaly images as supervised data, the anomaly detection model can be easily trained without complex adjustments and successfully detects anomalies.

\end{abstract}


\section{Introduction}
The railroad is one of Japan's most widely known means of transportation, and it necessitates high quality maintenance to ensure daily stability. However, because inspecting inputs requires a considerable amount of work and there is a risk of oversight, methods for automatic railroad inspection are being developed\cite{Chenariyan2019}. JR East Japan Group, Japan's largest railroad company, is working on smart maintenance that conforms to the state of the equipment and vehicles\cite{jre_revo2027}; however, because maintenance is performed before anomalies occur owing to meticulous quality management, much anomaly data is not available. The visual inspection of the underfloor equipment of rolling stock, which is the subject of this research, is performed by human eyes, but anomaly detection has been studied to automate part of this visual inspection\cite{jrerolling}.
They were successful in detecting anomalies in the study by comparing the luminance of normal and anomalous images. However, it is difficult to apply anomaly detection per supervised learning, which necessitates a large number of anomaly images for the supervised data.
In machine learning, securing sufficient data is essential. There are numerous data expansion methods via creating machine learning data via amplifying a small amount of data, which are summarized in a previous study\cite{shorten2019survey}. As reported in Ref.\cite{Frid_Adar_2018}, the amount of data was increased by generating images of liver lesions using deep convolutional GAN (DCGAN)\cite{radford2016unsupervised}, to classify the illness images. Moreover, CycleGAN was used to amplify a small number of facial image data\cite{zhu2017data, zhu2020unpaired}. The method for amplifying anomaly data described above, however, uses GAN. As a result, because supervised learning requires different anomaly data for learning data and test data, it cannot be used when only one set of anomaly data is available. Alternatively, AnoGAN is an unsupervised learning method that does not use anomaly data\cite{schlegl2017unsupervised}. AnoGAN can generate anomaly detection models without the use of anomaly data, but it has some limitations, such as a high computational cost. Therefore, f-AnoGAN was developed afterward\cite{SCHLEGL201930}, which greatly reduced the computational cost. In generative anomaly detection, a model trained only with normal images cannot reproduce anomaly regions, and errors occur during reconstruction. Recently, extensive research has been done to better understand the range of reconstruction of normal images. Using multiple hypotheses, the normal image's range was clearly captured\cite{nguyen2019anomaly}. According to previous studies, the range of the normal image is defined as a Gaussian distribution\cite{wang2020}; a model was built using double autoencoders GAN\cite{tang2020}; and a rough reconstruction model was combined with a detailed reconstruction model\cite{liu2021unsupervised}. The model accuracy has been improved by reducing the influence of the background\cite{Kimura_2020_WACV, song2021attention}. Another study increased the accuracy through training model with a small number of anomalous images\cite{lee2021deep}. Furthermore, the accuracy was improved by generating abnormal images based on normal images and using them for training\cite{wang2018anomaly, Zavrtanik_2021_ICCV}.
The anomaly detection using only normal images as described above is very important in the railroad field.
As a substantial amount of anomaly imagery is not available in the railroad maintenance field, many railroad corporations and organizations are exploring it. To address cases where there is a lot of data but a small percentage of anomaly images, luminance was improved by detecting anomaly data via semisupervised learning and adding it to the training data\cite{Hajizadeh2016}. Simultaneously, a model was developed for detecting anomalies in an overhead wire using a DCGAN discriminatory mechanism, as well as a model for detecting anomalies in railroad equipment using unsupervised learning that does not use anomaly data\cite{Lyu2019}. Moreover,, models were used that also include encoders that convert image data to vector spaces, to create anomaly detection models for electric equipment\cite{Lu2020,zhao2021}. These models do not use anomaly images, but there was still room for improvement in terms of luminance. A method for detecting anomalies was proposed by
 overlaying rail images of the same area and detecting the differences\cite{Gao2020,Guo2020}. This method can be used if there is only one image as a sample per shot or area, but it required machine learning using anomaly images to prevent false positives that could likely occur.
Alternatively, there is a method for adjusting for a lack of anomaly data in industrial and robot detection by creating anomaly images with three-dimensional (3D) computer graphics (CG) software and using them as supervised images. It is possible to generate anomaly images that do not exist using this method, but there will be issues with color and texture differences between CG images and actual photos, as well as being unable to distinguish actual images even when training with CG images. CG simulation images were used to train the object detection model of robotic controls, and they enable recognition of “one of many variations” for actual images by changing the rendering settings, generating images in various brightness, and colors, and using them as supervised images\cite{tobin2017domain}. Further, images were generated by adjusting the texture settings of the objects to detect, approximate the actual texture, and use images from the actual site for the background portions\cite{KUDO2019832}. Based on this research, we can see that “object detection” using CG for the supervised data was successful, but it does not refer to “anomaly detection” that detects small changes in images. Anomaly detection is a more difficult task than object detection because the precision changes depending on small differences in the texture and luminance when compared to object detection.
In this study, we used GANs on CG to generate anomaly images that were infinitely close to the real images shot with an external inspection device and then used those as supervised data to detect anomalies.

\section{Data}
\subsection{Vehicle side imagery and piping cocks}
\begin{figure}[ht]
  \includegraphics[width=\columnwidth]{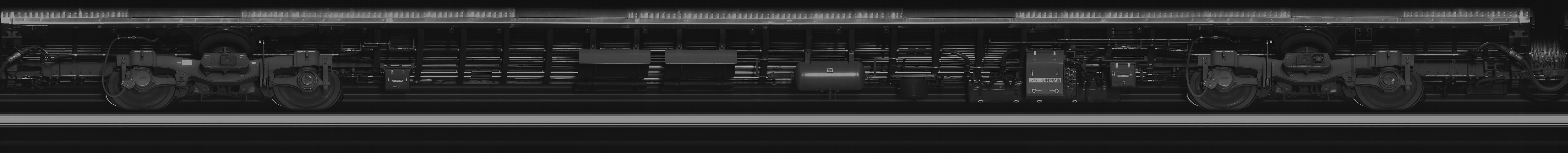}
  \caption{Image captured with an external vehicle inspection device)}
    \label{fig:long_image}
\end{figure}

\begin{figure}[ht]
  \begin{center}
  \includegraphics[width= 0.6\columnwidth]{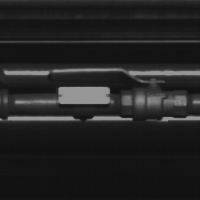}
  \caption{Extracted image of piping cock area Anomaly images}
    \label{fig:cock}
  \end{center}
\end{figure}

The images of vehicle sides captured by external vehicle inspection devices are the focus of this study. Fig.\ref{fig:long_image} depicts an image taken with an external vehicle inspection device. Because the resolution of this image is very high (20,501 × 2,048 pixels) and it was taken over a large area–the entire vehicle. Therefore, it would be difficult to apply machine learning anomaly detection. As a result, we investigated whether anomalies in piping cocks could be detected. A trimmed image of 200 × 200 pixels with the piping cocks and their surrounding area extracted is shown in Fig.\ref{fig:cock}.
The piping cock is considered to be in normal condition in Fig.\ref{fig:cock}. If the piping cock angle differs from that in Fig.\ref{fig:cock}, it is determined to be an anomaly.

\begin{figure}[ht]
  \includegraphics[width=\columnwidth]{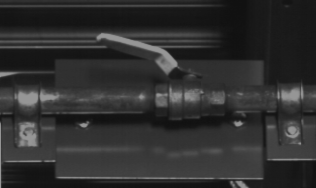}
  \caption{Anomaly image}
    \label{fig:anomaly_image}
\end{figure}

The anomaly images required for the verification of the anomaly detection model were provided by the East Japan Railway Company. The anomaly images are shown in Fig.\ref{fig:anomaly_image}.

\subsection{Data flow}
\begin{figure}[ht]
  \includegraphics[width=\columnwidth]{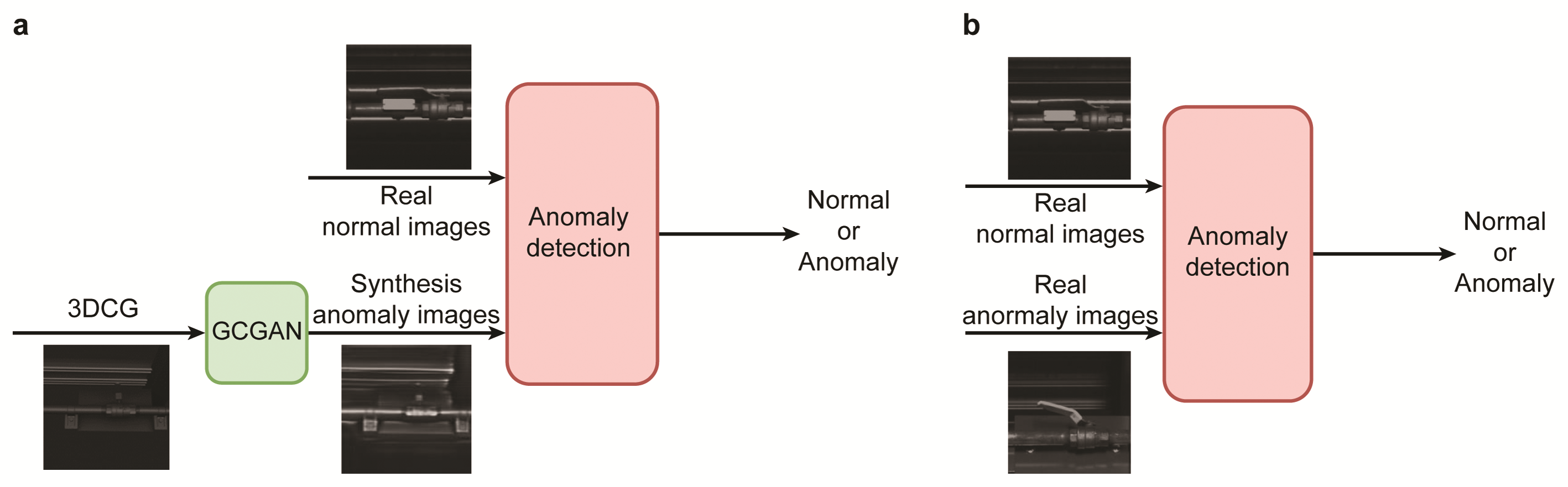}
  \caption{Data flow: (a) flow of training anomaly detection model (b) flow of testing}
    \label{fig:data_flow}
\end{figure}

Fig.\ref{fig:data_flow} depicts the data flow of this study. We used an image of the actual device for training and testing for the normal image. Meanwhile, the data used for training and testing for the anomaly image is different. We used data created in 3D CG and converted it using the image conversion model for machine learning for the training data. The test data consisted of anomaly images provided by East Japan Railway Company. The details of the training image generation method and the number of images used are described in \ref{subsec:data_set}.

\subsection{CG imagery}
\begin{figure}[ht]
  \includegraphics[width=\columnwidth]{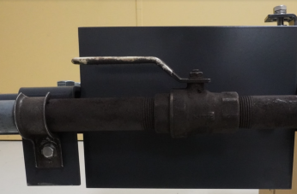}
  \caption{Actual object of the anomaly simulator}
    \label{fig:anomaly_simulator}
\end{figure}

\begin{figure}[ht]
  \includegraphics[width=\columnwidth]{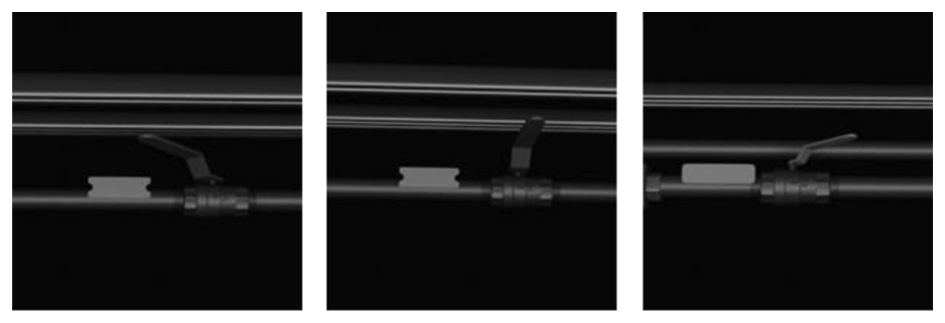}
  \caption{Sample of computer graphics anomaly in piping cock}
    \label{fig:sample_cg}
\end{figure}

Based on photos of the anomaly simulator, we created the CG to use as supervised data. Fig.\ref{fig:anomaly_simulator}  depicts a representation of the anomaly simulator. We generated 3D geometric data that replicated the geometry of the photo(s) as closely as possible. To create anomaly CG, we used 3D rendering software based on 3D geometric data and exported an image rotated at various angles. Fig.\ref{fig:sample_cg} depicts an example of the CG anomaly. In comparing Fig.\ref{fig:cock} and Fig.\ref{fig:sample_cg}, the surface of the CG image is more even and does not have the color shading that is in the actual image.


\section{Proposed Method}
\subsection{Style conversion via GAN}
\begin{figure}[ht]
  \includegraphics[width=\columnwidth]{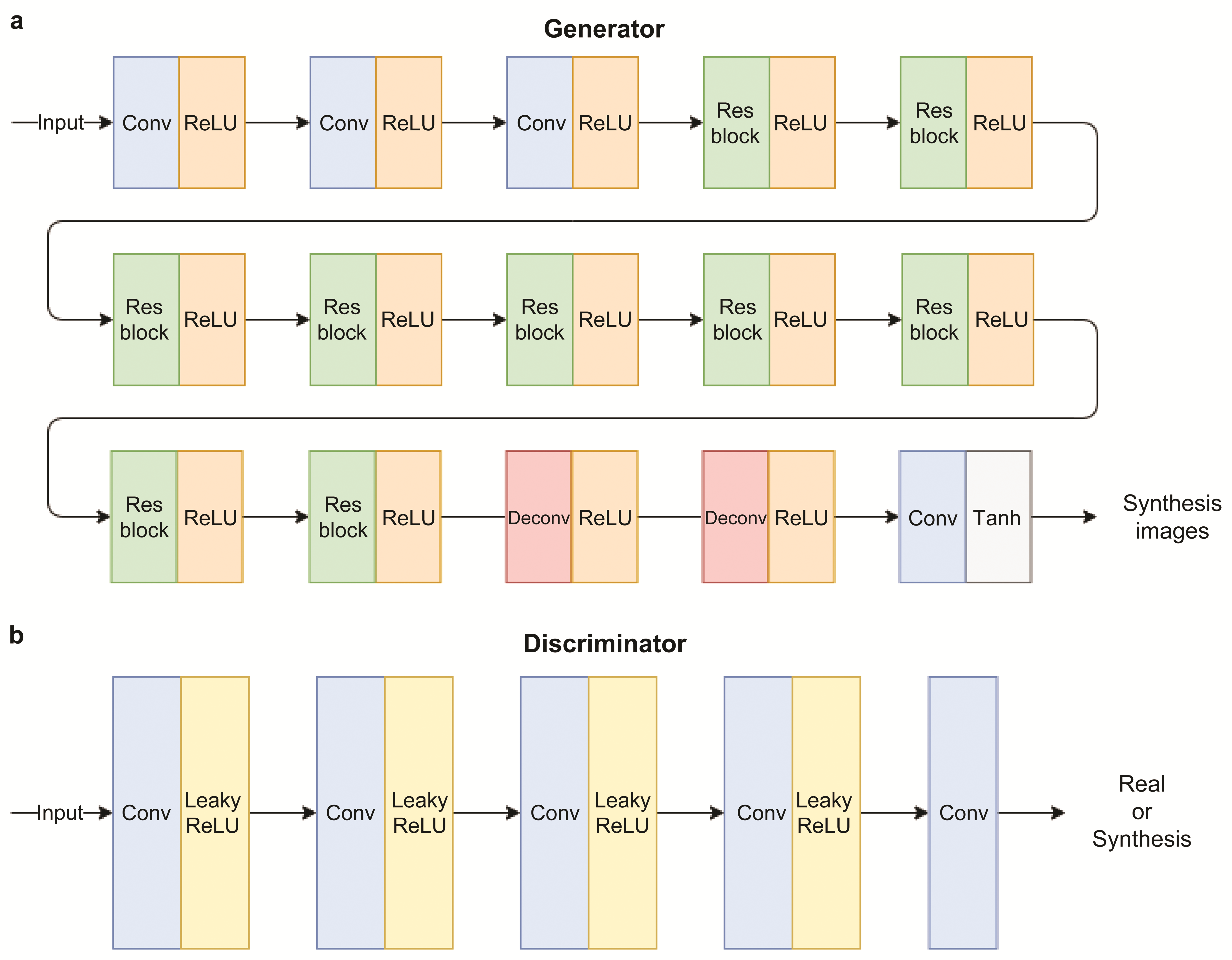}
  \caption{Network architecture of geometry-consistency generative adversarial networks}
    \label{fig:GCGAN}
\end{figure}

To make the CG image closer to the actual image, we performed style conversion using machine learning. In this study, we used geometry-consistency GAN (GCGAN)\cite{fu2018geometryconsistent}, believing that converting the color and texture while maintaining the geometry was necessary. In GCGAN, we train so that “a normal converted image” and “a rotated image that is converted and then reverted to its original rotation” match. This helps us to maintain precise geometry while converting the color and pattern. The GCGAN we used was the model published by GitHub\cite{gcgangithub}. The architecture is shown in Fig.\ref{fig:GCGAN}.

\begin{figure}[ht]
  \includegraphics[width=\columnwidth]{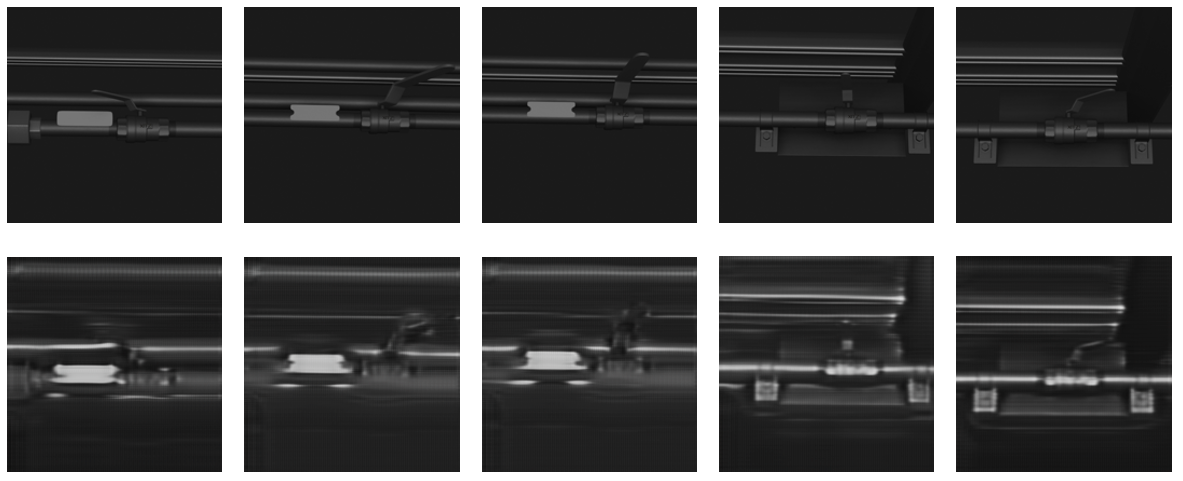}
  \caption{Results of converting computer graphics with geometry-consistency generative adversarial networks}
    \label{fig:result_image_gcgan}
\end{figure}

As learning data, we used 600 actual image data and 600 CG data. The dropout layer was disabled, the input size was 200 × 200, the batch size was 12, and the transformations could be flipped vertically or rotated 90°, so we tried both and used those flipped vertically. The IdentityMappingLoss parameter was set to 0.5, which determines the percentage of the original image’s color and structure that is retained. We trained the model for 400 epochs without changing the learning rate, then for 200 epochs while decreasing the learning rate to 0, and used the model with the best quality from the 200th epoch. We input CG anomaly data and generated 1,000 anomaly data against the model that had been trained. Fig.\ref{fig:result_image_gcgan} shows an example of one of the generated images. The rotated piping cock and plate are reproduced in the same position, and it can be confirmed that the color shading and texture are changed.

\subsection{Data set}
\label{subsec:data_set}
To compare the results of anomaly detection precision trained using CG and anomaly images that had style conversion with GCGAN, we created an anomaly detection model from two patterns of data sets. The details of the data are shown in Table\ref{tab:data_set}.

\begin{table}[ht]
\begin{tabularx}{\columnwidth}{|X|X|X|X|X|}
    \hline
    \textbf{} & \multicolumn{2}{c|}{\textbf{CG Training Model}}  & \multicolumn{2}{c|}{\textbf{GCGAN Training Model}} \\
    \hline  & Training & Data & training & Data \\
    \hline  Actual Image-Normal & 600 & 600 & 600 & 600 \\
    \hline  CG Anomaly & 600 & 0 & 0 & 0 \\
    \hline  GCGAN-Anomaly & 0 & 0 & 600 & 0 \\
    \hline  Actual Image-Anomaly & 0 & 1 & 0 & 1 \\
    \hline 
\end{tabularx}
\caption{Data set of anomaly detection model}
\label{tab:data_set}
\end{table}

\subsection{Anomaly detection model using ResNet}
\begin{figure}[ht]
  \includegraphics[width=\columnwidth]{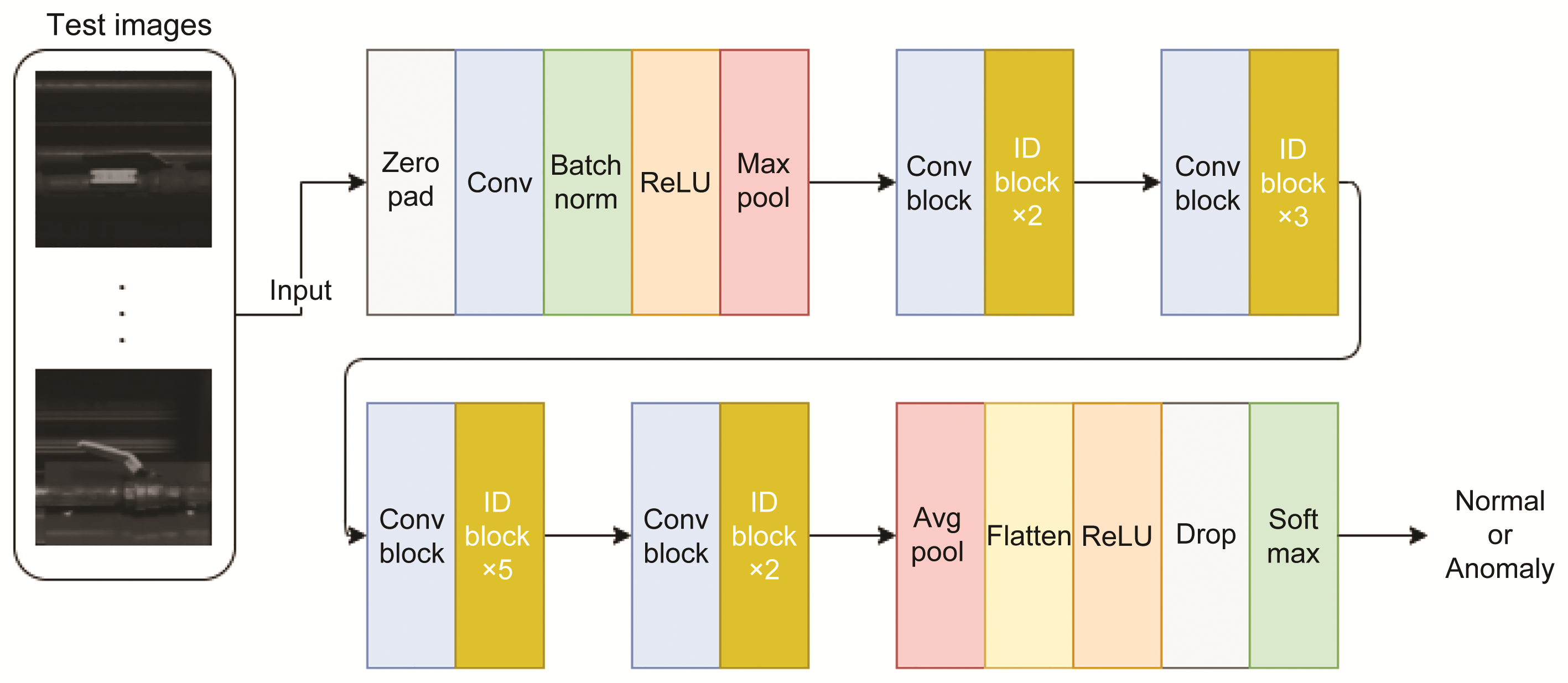}
  \caption{Structure of anomaly detection model}
    \label{fig:resnet_model}
\end{figure}

To test whether anomaly detection is possible with anomaly images generated using GCGAN as supervised data, we implemented transfer learning based on ResNet.\cite{he2015deep} The structure is shown in Fig.\ref{fig:resnet_model}. The ResNet50 model of Tensorflow was used, and the configuration is shown in Fig.\ref{fig:resnet_model}. We used a learning rate of $ 1.0 * 10^{-5} $, a training frequency of 50 epochs, a momentum of 0.9, a batch size of 32, and an image size scaled to 224. Two patterns, CG and GCGAN, were used as the anomaly images for training, and both were trained using the same parameters.

\section{Results and Discussion}
\subsection{Results}
\begin{figure}[ht]
  \includegraphics[width=\columnwidth]{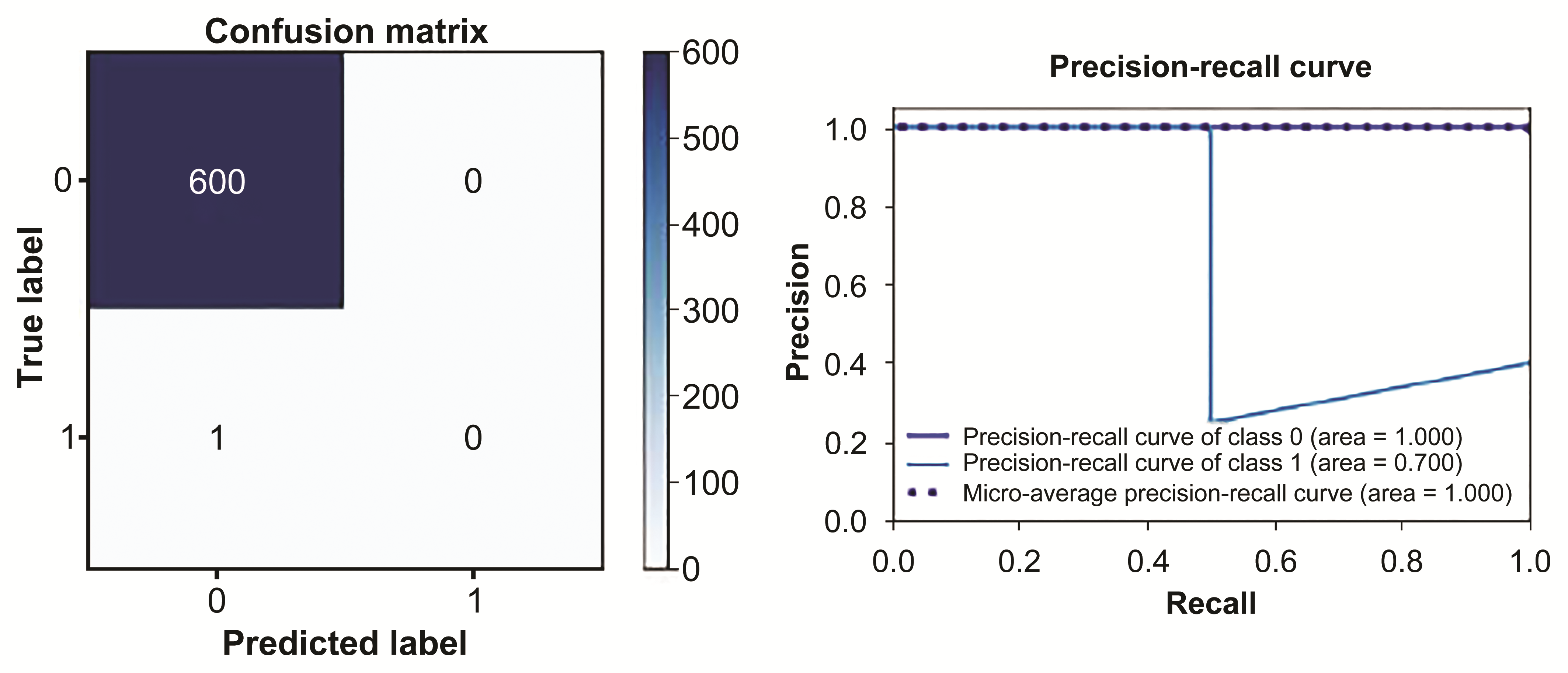}
  \caption{Detection precision of computer graphics training model (0: normal, 1: anomaly)}
    \label{fig:resut_cg}
\end{figure}

\begin{figure}[ht]
  \includegraphics[width=\columnwidth]{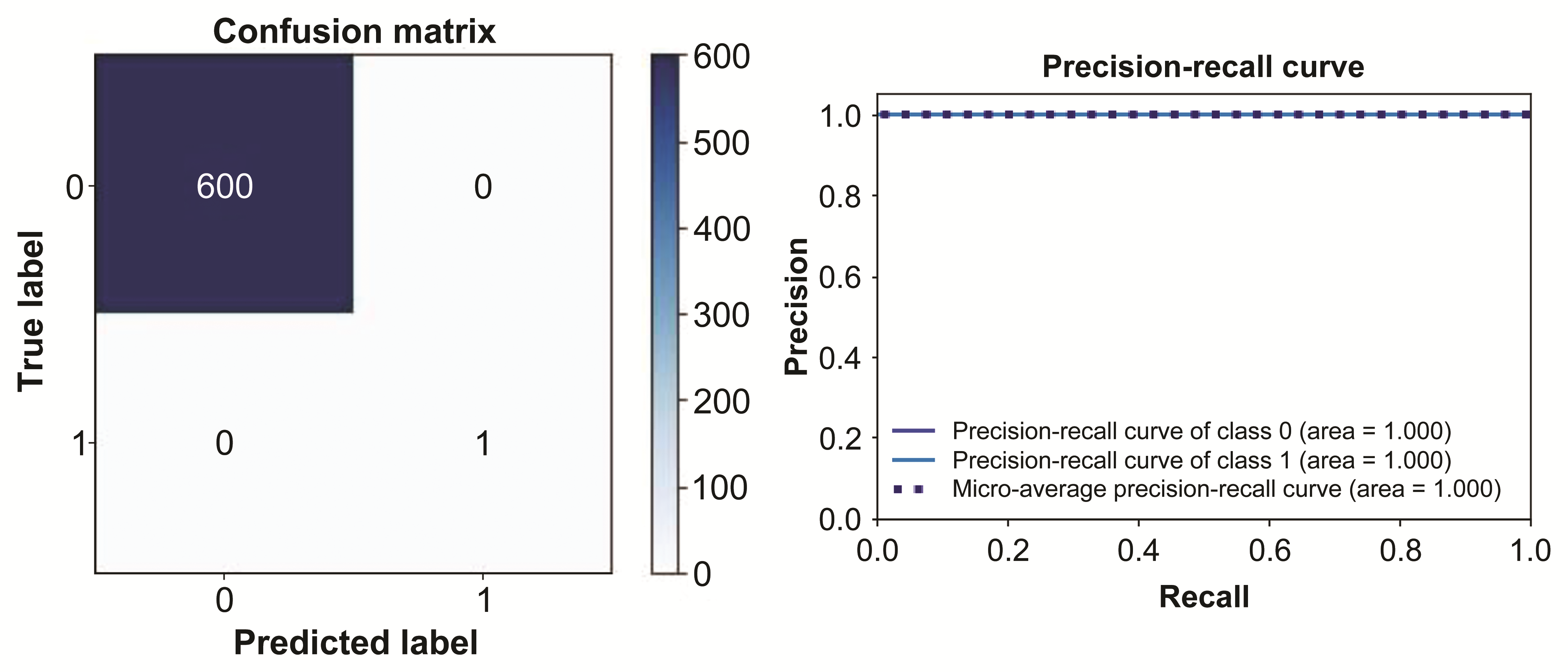}
  \caption{Detection Precision of GCGAN Training Model (0: normal, 1: anomaly)}
    \label{fig:resut_gcgan}
\end{figure}

Fig.\ref{fig:resut_cg} depicts the outcomes of anomaly detection using only CG as supervised data. We discovered that the anomaly images could not be distinguished at all and that improving the precision with CG anomaly images was difficult. Fig.\ref{fig:resut_gcgan}  then displays the results of GCGAN training model detection. The GCGAN training model was effective in detecting anomalous images and increasing accuracy.


\subsection{Discussion}
\begin{figure}[ht]
  \includegraphics[width=\columnwidth]{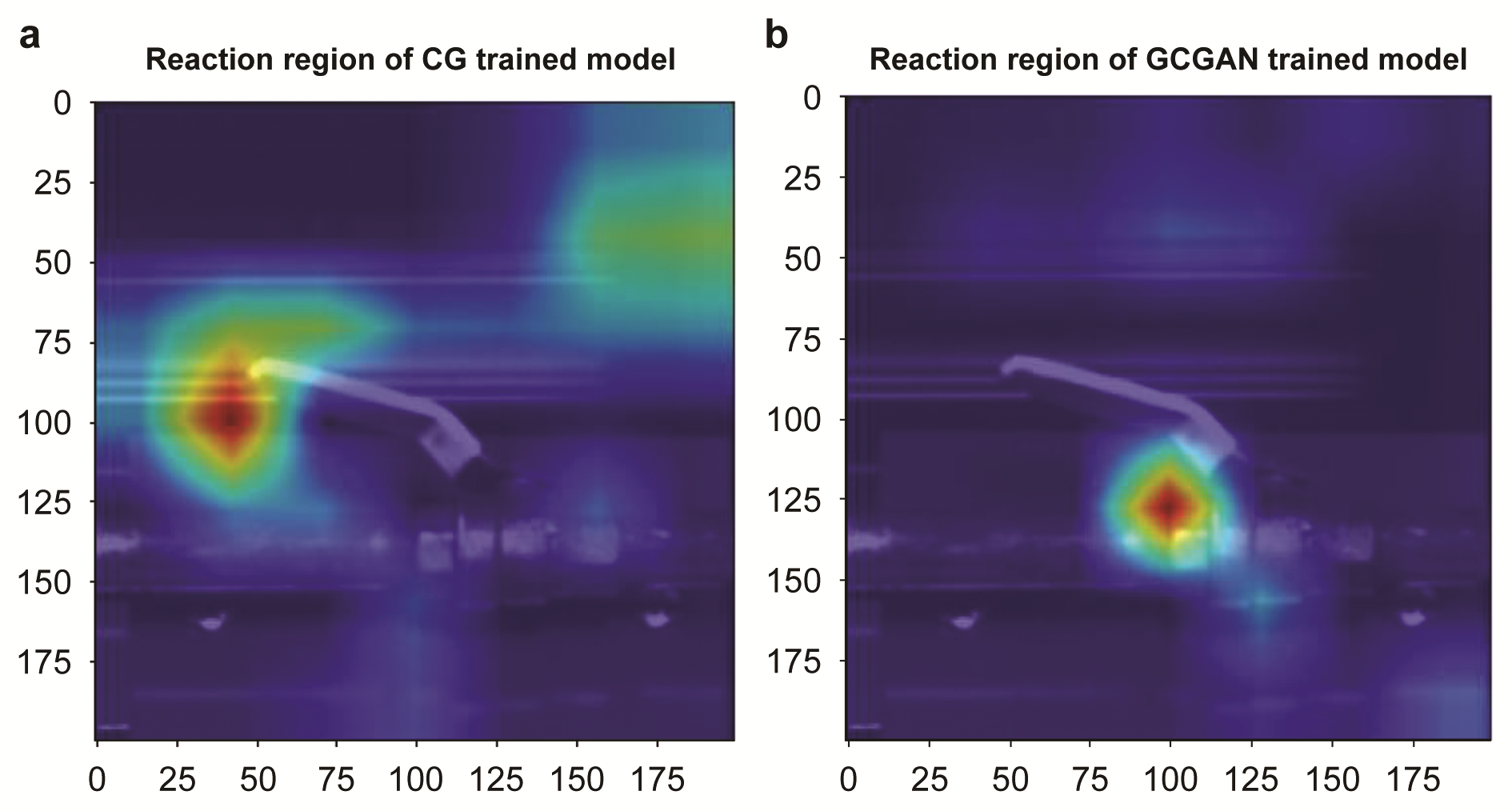}
  \caption{Result of Grad-CAM (a)Reaction region of CG trained model (b)Reaction region of GCGAN trained model}
    \label{fig:heat_map}
\end{figure}

There was a difference in the detectability of anomaly images when training with CG and when using CG converted using GCGAN. Moreover, whether this is sufficient to focus on the position of the cock is unclear. As a result, we used Grad-CAM to visualize the areas that the anomaly detection model considers when making decisions\cite{Selvaraju_2019}. The results of Grad-CAM when an anomaly image was used are shown in Fig.\ref{fig:heat_map}. The model's reaction is spread across the entire image due to the use of CG. The model based on GCGAN, however, focuses on the handle positions. This indicates that GCGAN's texture transformation changed the detection model's focus to the image's structure.

\section{Conclusions}
In this research, we verified that anomaly detection is possible using fictional anomaly images generated by combining CG and GCGAN as supervised data. A common issue in railroad maintenance is the scarcity of anomaly data. When a specific number of anomaly patterns can be obtained from a small amount of data (for example, when there are several dozen to several hundred anomaly data), there is existing research that allows for an increase in data through data expansion. Furthermore, even if there was a way to build a model using unsupervised learning without using any anomaly data, and the images were normal in the unsupervised learning, there would be more false positives owing to environmental changes, causing precision issues. In this research, we were able to identify anomaly patterns by generating fictional anomaly images and succeeded in detecting anomalies without increasing the number of false positives. In this study, efforts in creating models specific for this research using 3D CG software were required, but we believe that, in the future, we could collect learning data with a simulator, as has been done in related research on robot recognition. In the future, we will expand on this research, address the issue of scarce anomaly data that is shared across the railroad industry, and aspire to implement automatic anomaly detection in society.

\section*{Acknowledgement}
We wish to thank East Japan Railway for supplying sample data. We would like also to thank colleagues in Technical Research Center in JR East Information Systems for useful discussions.

\bibliography{main}

\end{document}